\documentclass[a4paper]{article}

\usepackage{INTERSPEECH2019}
\usepackage{subfigure}

\title{Multi-reference Tacotron by Intercross Training for Style Disentangling, Transfer and Control in Speech Synthesis}
\name{Yanyao Bian, Changbin Chen, Yongguo Kang, Zhenglin Pan}
\address{
  Baidu Speech Department\\
  Baidu Inc. Baidu Technology Park, Beijing, 100193, China}
\email{\{bianyanyao, chenchangbin, kangyongguo, panzhenglin\}@baidu.com}

\begin{document}

\maketitle
\begin{abstract}
Speech style control and transfer techniques aim to enrich the diversity and expressiveness of synthesized speech. Existing approaches model all speech styles into one representation, lacking the abilty to control a specific speech feature independently. To address this issue, we introduce a novel multi-reference structure to Tacotron and propose intercross training approach, which together ensure that each sub-encoder of the multi-reference encoder independently disentangles and controls a specific style. Experimental results show that our model is able to control and transfer desired speech styles individually.
\end{abstract}

\section{Introduction}\label{sec:introduction}

Recent years neural text-to-speech (TTS) has been developing rapidly and has become an indispensable part of our daily life. End-to-end-TTS (E2E) models, which generate speech without handcraft linguistic features, are one of the most attractive TTS techniques and have achieved great progresses in voice quality~\cite{shen2018natural, ping2017deep, li2018close}. However, it remains a challenge for E2E models to control speech styles, such as speaker, emotion, prosody, etc., which are essential for expressive and diverse voice generation. 

Currently there are mainly two ways to control and transfer speech styles. The supervised way, which takes attributes' id as an additional model input, has been proved effective in multi-speaker TTS~\cite{gibiansky2017deep}. However, this approach can hardly deal with more complex styles like emotion and prosody, because there are no objective measurements to annotate these styles. The unsupervised way, which combines E2E models and reference encoders into Autoencoder (AE)~\cite{skerry2018towards} or Variational Autoencoder (VAE)~\cite{akuzawa2018expressive, zhang2018learning, hsu2018hierarchical}, gets closer results. Theoretically, the unsupervised E2E models can model any complex styles in a continuous latent space, so that one can control and transfer style by manipulating the latent variables.

However, most of the previous researches model all speech styles into one style representation, which contains too much interfering information to be robust and interpretable. When conducting style transfer, one has to transfer all styles whether desired or not, which may not fit the contexts thus hurts generalization. When conducting style control, one can hardly confirm the relationship between the styles and the coefficients of each dimension of the style representations. Therefore, a natural question then arises: can we control different styles independently in an interpretable way? 

To address the above issues, (1) we introduced a multi-reference encoder to GST-Tacotron~\cite{wang2018style}, a state-of-the-art E2E model for style disentangling, control and transfer, to model multiple styles simultaneously. (2) We introduced intercross training (IT) to extract and separate different classes of speech styles. Several experiments are conducted to show the proposed methods' feasibility.


The rest of the paper is organized as follows. Section~\ref{sec:model_and_metrics} introduces multi-reference structure, intercross training, auxiliary tasks and inference procedures. Section~\ref{sec:experiments} presents experimental results. Section~\ref{sec:conclusions} gives a conclusion of this paper. 

\section{Model and metrics}\label{sec:model_and_metrics}



\subsection{Model structure}\label{sec:proposed_model}

\begin{figure}[t]
  \centering
  \subfigure[Overall structure]{
    \label{fig:structure.a}
    \includegraphics[width=1\linewidth, height=1.4in]{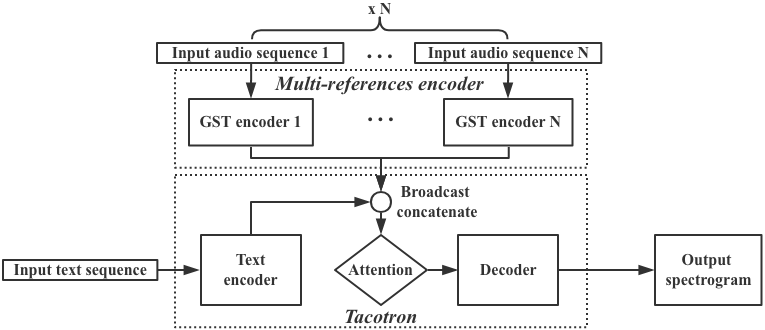}}
  \subfigure[Multi-reference encoder]{
    \label{fig:structure.b}
    \includegraphics[width=1\linewidth, height=1.7in]{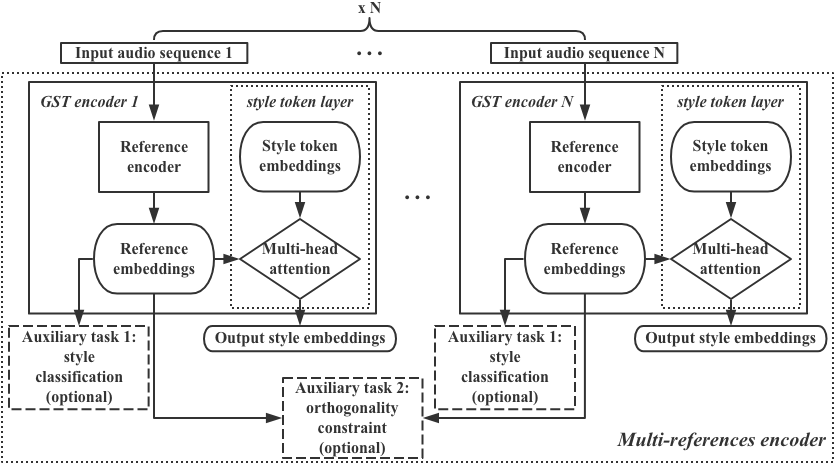}}
  \caption{Structure of the proposed model.}
  \label{fig:structure}
\end{figure}

Our model is based on GST-Tacotron. Figure~\ref{fig:structure.a} shows its overall structure, which mainly consists of two parts: Tacotron and multi-reference encoder. Figure~\ref{fig:structure.b} shows the detailed structure of multi-reference encoders, which consists of $N$ GST sub-encoders, abbreviated as sub-encoders. The Tacotron and GST-encoder in this paper have no differences from those in \cite{wang2018style}. We use Griffin-Lim~\cite{griffin1984signal} for fast experiment and use WaveNet~\cite{oord2016wavenet} to generate demos for better audio fidelity.



For each sub-encoder, the reference encoder takes a reference audio (or acoustic parameters) as input and outputs the \textit{reference embeddings}. Then a multi-head attention is conducted between reference embeddings and the \textit{style token embeddings} to generate the \textit{style embeddings}. Finally the style embeddings of each sub-encoder are broadcast and concatenated to the outputs of text encoder for standard Tacotron decoding. In addition, we add two optional auxiliary tasks in multi-reference encoder for training our models, which will be discussed later.

\subsection{Definitions}\label{sec:definitions}
For better explaining how to control multiple styles, we firstly introduce the definitions of styles in this paper.

The \textbf{\textit{style class}} is the abstract description of styles which we want to disentangle and control. In our experiments, we evaluate 3 style classes: speaker, emotion, prosody. However, the proposed method is not constrained on these style classes.

The \textbf{\textit{style instance}} refers to the instances of style classes. In this paper, speaker class is instantiated as 300 different speakers; emotion class is instantiated as happy, sad, angry, fear, confuse, surprise and neutral; prosody class is instantiated as news, story, radio, poetry and call-center.

In this paper, we use $X_{i_{1}i_{2}...i_{N},j}$ to represent an audio (or acoustic parameters) with $N$ style classes to discuss, each of which instantiated as $i_1, i_2, ..., i_N$, and with text content $j$. Furthermore, we introduce a wildcard ``*'' to represent ``any''. For example, $X_{*i_n*,j}$ represents an audio with style instance $i_n$ , text content $j$ and a number of arbitrary other style instances.


Each sub-encoder in the multi-reference encoder is designed to model only one style class. Formally, we want the following equation (\ref{eq1}) to be maintained in each sub-encoder.

\begin{equation}
\begin{split}
  Q_{\phi_n}(S_{i_n}{\mid}&X_{*i_n*,j}) = Q_{\phi_n}(S_{i_n}{\mid}X_{*i_n*,k}) \\
            &for\ n = 1, 2, ..., N
\end{split}
\label{eq1}
\end{equation}

Where $S_{i_n}$ represents style instance $i_n$; $\phi_n$ represents the parameters of n-th sub-encoder. 

This equation is saying that the n-th sub-encoder provides the same posterior distribution given audios with the same style instance of the n-th style class, regardless of other style classes and text contents. For example, in a two-reference model, we can have 1-st sub-encoder focus on modeling speaker style class and generating speaker embeddings, while utilizing the 2-nd sub-encoder to deal with prosody style class. Consequently, we can control different style classes independently by manipulating their own style embeddings.



\subsection{Intercross training}\label{sec:intercross}

Consider the simplest case $N = 1$. If we pick a pair of audios $\{X_{i_1,j}, X_{i_1,k}\}$ with the same style instance $i_1$ and text contents $j \ne k$, simultaneously minimizing following reconstruction loss functions to zero guarantees the equation (\ref{eq1}) holds: 

\begin{equation}
\begin{split}
  \mathcal L_{ORG}&(X_{i_1,j}, X_{i_1,k}; \theta, \phi_1) = \\
                    & - E_{Q_{\phi_1}(S_{i_1}{\mid}X_{i_1,j})}[\log{P_\theta(X_{i_1,j}{\mid}S_{i_1},c_j)}] \\
                    & - E_{Q_{\phi_1}(S_{i_1}{\mid}X_{i_1,k})}[\log{P_\theta(X_{i_1,k}{\mid}S_{i_1},c_k)}]
\end{split}
  \label{eq2}
\end{equation}

\begin{equation}
\begin{split}
  \mathcal L_{IT}&(X_{i_1,j}, X_{i_1,k}; \theta, \phi_1) = \\
                    & - E_{Q_{\phi_1}(S_{i_1}{\mid}X_{i_1,k})}[\log{P_\theta(X_{i_1,j}{\mid}S_{i_1},c_j)}] \\
                    & - E_{Q_{\phi_1}(S_{i_1}{\mid}X_{i_1,j})}[\log{P_\theta(X_{i_1,k}{\mid}S_{i_1},c_k)}]
\end{split}
  \label{eq3}
\end{equation}

Where $\theta$ represents the parameters of Tacotron. Equation (\ref{eq2}) is the original object whose reference and target are the same, and equation (\ref{eq3}) are called intercross object as we exchange the $Q_{\phi}(S{\mid}X)$ when doing reconstruction. After fully optimizing the two objects, we see $\mathcal L_{ORG} = \mathcal L_{IT}$ ideally. The posterior distributions of different audios $P_\theta(X{\mid}S,c)$ are different, hence the exchanged part must be equal, then equation (\ref{eq1}) holds. 

Notice that equation (\ref{eq2}) can be included in equation (\ref{eq3}) by removing the constraint $j \ne k$. Furthermore, with the assumption that a fully training procedure would have enumerated all reference-target combinations, we could optimize the two parts of equation (\ref{eq3}) separately in different training steps as an approximation. Therefore, the final objective function of $N = 1$ intercross training comes to the following equation (\ref{eq.final1}).

\begin{equation}
\begin{split}
  \mathcal L_{IT} = -E_{Q_{\phi_1}(S_{i_1}{\mid}X_{i_1,*})}[\log{P_\theta(X_{i_1,j}{\mid}S_{i_1},c_j)}]
\end{split}
  \label{eq.final1}
\end{equation}

To minimize equation (\ref{eq.final1}), we randomly pick two audios with replacement from a group of audios $\mathcal G = \{X_{i_1,*}\}$, where audios in $\mathcal G$ share the same style instance $i_1$. Then we use one of them as reference and the other as target to train the model.




For $N > 1$ cases, equation (\ref{eq.final1}) can be generalized to equation (\ref{eq4}) according to the multi-reference structure, which takes $N$ styles from $N$ reference inputs to construct the target.


\begin{equation}
\begin{split}
  \mathcal L_{IT} = & -E_{\prod_{n=1}^N Q_{\phi_n}(S_{i_n}{\mid}X_{*i_n*,*})} \\
                    &[\log{P_\theta(X_{i_1i_2...i_N,j}{\mid}S_{i_1i_2...i_N},c_j)}]
\end{split}
  \label{eq4}
\end{equation}

To minimize equation (\ref{eq4}), we firstly randomly pick a target $T = X_{i_{1}i_{2}...i_{N},j}$. Then we randomly pick $N$ references $R = \{X_{i_{1}*,*}, X_{*i_{2}*,*}, ..., X_{*i_{N},*}\}$ with replacement, whose n-th element $R_n = X_{*i_{n}*, *}$ shares the same style instance $i_n$ with $T$. Finally we construct the references-target pair (R, T) as a sample to train the model.

We have realized that the concept of intercross training has been mentioned before in \cite{sun2017learning}. However, they did not discuss its general form for multi-style disentangling in AE structures.

\subsection{Auxiliary tasks}

In our experiments, our model did not converge well because it was confused by the meanings of multi-reference encoder style embeddings. To address this issue, we introduce a style classification task $\mathcal L_{classification}$ over the reference embeddings. Intuitively, the style classification task ensures that the desired style information has been already modeled in the reference embeddings, and it will be easier for intercross training to squeeze out other redundancies.

Inspired by \cite{bousmalis2016domain}, we further introduce an orthogonality constraint shown in equation (\ref{eq5}) to strengthen the independence of the style embeddings of each sub-encoder.

\begin{equation}
  \mathcal L_{orthogonality} = \sum_{ij}\lVert{H_i^TH_j}\rVert_{F}^2
  \label{eq5}
\end{equation}

Where $\lVert\cdot\rVert_{F}^2$ is the squared Frobenius norm and $H_i, H_j$ means the reference embeddings of the i-th and j-th sub-encoder, respectively. The final object comes to equation (\ref{eq6}):

\begin{equation}
  \mathcal L = \mathcal L_{IT} + \beta \mathcal L_{classification} + \gamma \mathcal L_{orthogonality}
  \label{eq6}
\end{equation}


In our experiments, we set $\beta$ to 1 and $\gamma$ to 0.02 without exploring. Notice that the auxiliary tasks are optional and only performed when $N > 1$. 

\subsection{Inference}\label{sec:inference}

\begin{figure}[th]
  \centering
  \includegraphics[width=\linewidth]{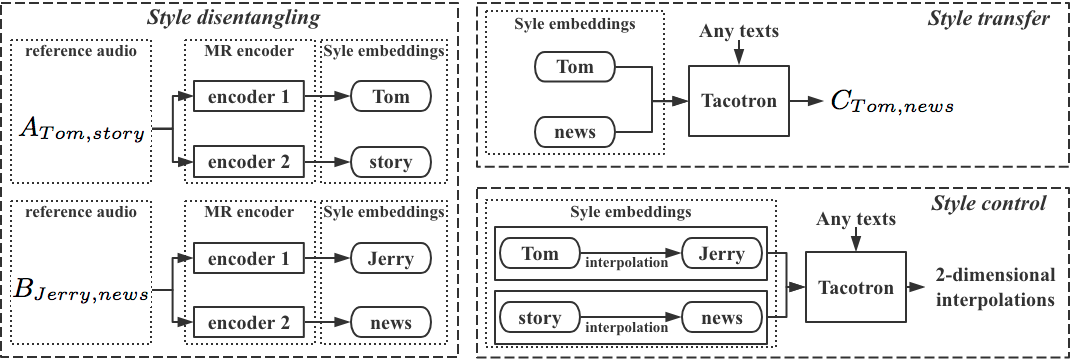}
  \caption{Inference procedure of style disentangling, transfer and control using the proposed model.}
  \label{fig:inference}
\end{figure}

Before experiments, we would like to introduce the inference procedures of the proposed model. We present an example of 2-references model with sub-encoder 1 to control speaker and sub-encoder 2 to control prosody. The inference procedures are shown in Figure~\ref{fig:inference}.

\textbf{\textit{Style disentangling}}. The fundamental of this paper described in equation (\ref{eq1}) and achieved by IT. Given a audios, each sub-encoder can model a specific style.

\textbf{\textit{Style transfer}}. Different disentangled style embeddings can be freely combined with each other to generate new voices.

\textbf{\textit{Style control}}. Styles can be controlled through linear interpolation between style embeddings of style instances in each encoder, formulated as equation (\ref{eq7}). 

\begin{equation}
  \begin{split}
    \boldsymbol {SE}_{to} = \boldsymbol {SE}_{from} + \alpha \cdot (\boldsymbol {SE}_{to} - \boldsymbol {SE}_{from})
  \end{split}
  \label{eq7}
\end{equation}

Where $\alpha \in [0, 1]$; $\boldsymbol {SE}$ refers to style embeddings.

\textbf{\textit{Random sampling}}. The proposed model is able to generate speech with random styles following equation (\ref{eq8}).

\begin{equation}
  \boldsymbol {SE}_{random} = \sum_{k=1}^K softmax(a_k) \cdot \boldsymbol {STE}_k
  \label{eq8}
\end{equation}

Where $\boldsymbol {STE}$ means the style token embeddings in a sub-encoder, shaped $K \times D$; $a_k \sim \mathcal U(0, 1)$. Speech generated in this way could be with any style instances of the style class controlled by current sub-encoder. 

\section{Experiments}\label{sec:experiments}

In this section, we firstly build a single-reference ($N = 1$) model to evaluate the performance of intercross training (IT). Then we build a multi-reference ($N = 2$) model to show the extendibility for controlling multiple styles.

In the following experiments, all speech data are recorded by BAIDU Speech Department and the sampling rate is 16 kHz. The model takes 80-dimensional mel-spectrogram with 50ms frame length and 12.5ms frame shift as reference input and output. Meanwhile, it is conditioned on Mandarin phonemes and tones. The model outputs 5 frames per decoding step. 

As the visualization in speech synthesis is often not enough for objective perception, we strongly encourage readers to listen to the audio samples provided on our demo page\footnote{https://ttsdemos.github.io/paper3.html}.

\subsection{Single-reference}\label{sec:single-reference}


In single-reference experiments, we mainly analyze the results of speaker style. The results of prosody style are presented in multi-reference experiments, and the results of emotion style can be found in our demo page.


We train models over an 110 hours corpus recorded by 300 different non-professional Chinese speakers including 178 females and 122 males, each of which recorded about 500 utterances. We use the original GST model (ORG) as comparison.

\subsubsection{Style disentangling}

\begin{figure}[th]
  \centering
  \subfigure[]{
  \label{fig:embedding_visualization.a}
  \begin{minipage}[b]{0.23\linewidth}
    \includegraphics[width=1\linewidth, height=0.7in]{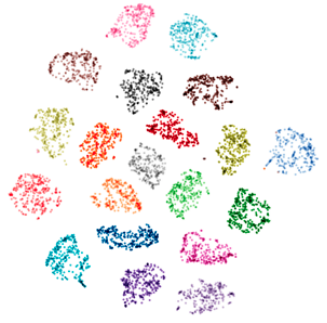}
  \end{minipage}}
  \subfigure[]{
  \label{fig:embedding_visualization.b}
  \begin{minipage}[b]{0.23\linewidth}
    \includegraphics[width=1\linewidth, height=0.7in]{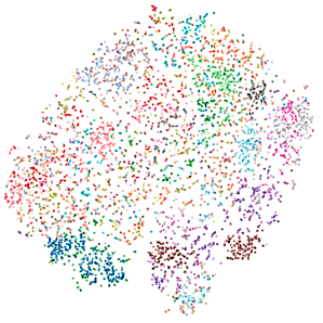}
  \end{minipage}}
  \subfigure[]{
  \label{fig:embedding_visualization.c}
  \begin{minipage}[b]{0.23\linewidth}
    \includegraphics[width=1\linewidth, height=0.7in]{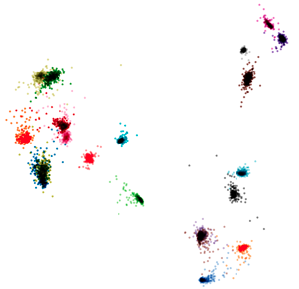}
  \end{minipage}}
  \subfigure[]{
  \label{fig:embedding_visualization.d}
  \begin{minipage}[b]{0.23\linewidth}
    \includegraphics[width=1\linewidth, height=0.7in]{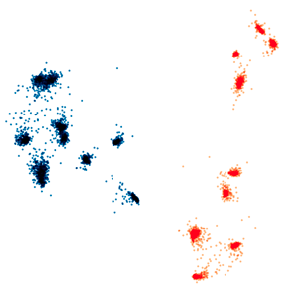}
  \end{minipage}}
  \caption{Style embeddings visualization. Pictures are colored by speakers if not specified. (a) IT-SE by t-SNE; (b) ORG-SE by t-SNE; (c) IT-SE by PCA. (d) IT-SE by PCA colored by genders.}
  \label{fig:embedding_visualization}
\end{figure}

We randomly chose 20 speakers including 10 females and 10 males and visualize the style embeddings of their about 10000 recordings using t-SNE~\cite{maaten2008visualizing} on Tensorboard shown in Figure~\ref{fig:embedding_visualization}.

Compared ORG-SE in Figure~\ref{fig:embedding_visualization.b}, we can clearly distinguish all 20 speakers with IT-SE in Figure~\ref{fig:embedding_visualization.a}. That is to say, unlike ORG-SE that contains lots of redundancies, IT-SE concentrate on speaker style class, which is consistent with equation (\ref{eq1}). Furthermore, we also visualize the same IT-SE by PCA, which is a linear dimension reduction method, in Figure~\ref{fig:embedding_visualization.c} and ~\ref{fig:embedding_visualization.d}. The results indicate that the IT-SE of different speakers and genders are linearly separable.

\subsubsection{Style transfer}

\begin{figure}[th]
  \centering
  \subfigure[Reference audios]{
  \label{fig:style_transfer_1.a}
  \begin{minipage}[b]{1\linewidth}
    \includegraphics[width=0.24\linewidth, height=0.4in]{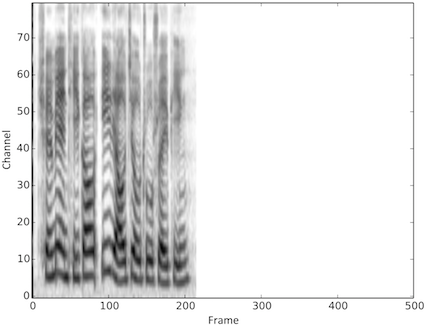}
    \includegraphics[width=0.24\linewidth, height=0.4in]{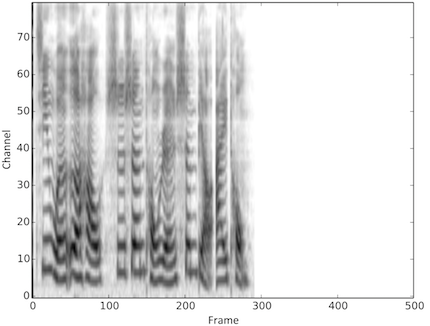}
    \includegraphics[width=0.24\linewidth, height=0.4in]{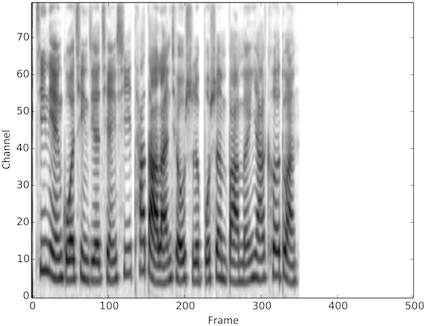}
    \includegraphics[width=0.24\linewidth, height=0.4in]{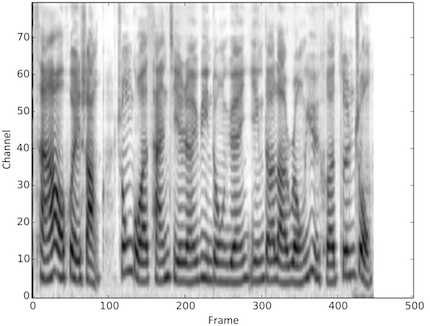}
  \end{minipage}}
  \subfigure[Outputs of ORG model]{
  \label{fig:style_transfer_1.b}
  \begin{minipage}[b]{1\linewidth}
    \includegraphics[width=0.24\linewidth, height=0.4in]{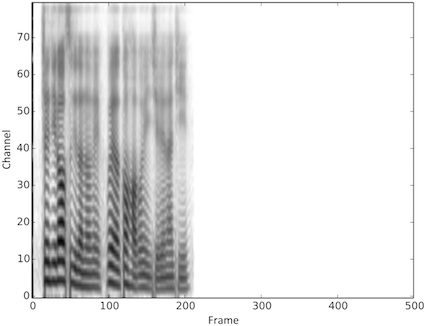}
    \includegraphics[width=0.24\linewidth, height=0.4in]{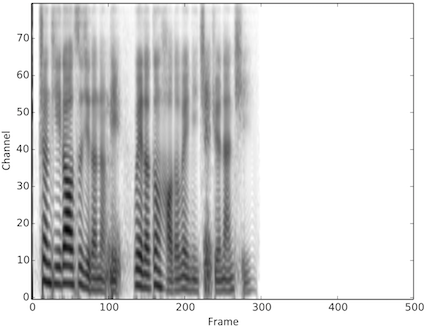}
    \includegraphics[width=0.24\linewidth, height=0.4in]{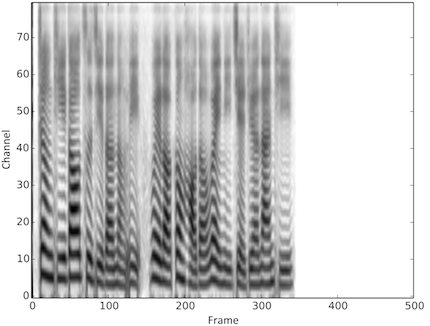}
    \includegraphics[width=0.24\linewidth, height=0.4in]{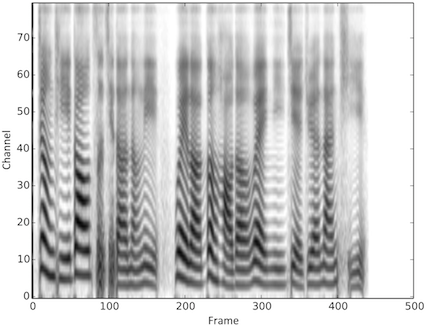}
  \end{minipage}}
  \subfigure[Outputs of IC model]{
  \label{fig:style_transfer_1.c}
  \begin{minipage}[b]{1\linewidth}
    \includegraphics[width=0.24\linewidth, height=0.4in]{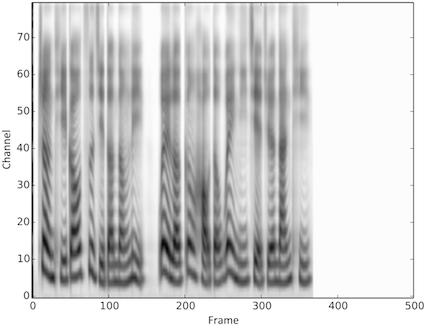}
    \includegraphics[width=0.24\linewidth, height=0.4in]{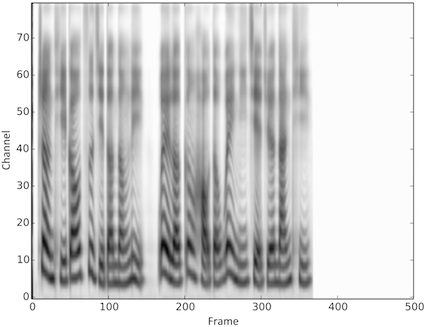}
    \includegraphics[width=0.24\linewidth, height=0.4in]{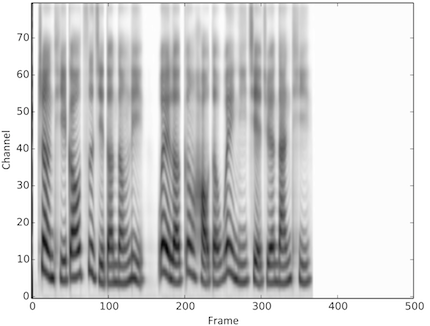}
    \includegraphics[width=0.24\linewidth, height=0.4in]{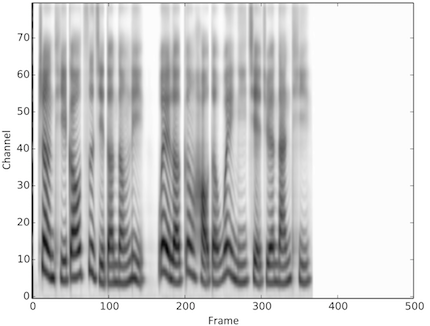}
  \end{minipage}}
  \caption{Non-parallel style transfer conditioned on same text and different reference audios from same speaker.}
  \label{fig:style_transfer_1}
\end{figure}

We conduct non-parallel style transfer by randomly choosing 4 various audios from a speaker as reference audios to synthesize a fixed unseen text. Figure~\ref{fig:style_transfer_1} shows the results.


We see that the output audios of the ORG model in Figure~\ref{fig:style_transfer_1.b} are with the lengths coinciding with reference audios instead of texts, and both naturalness and speaker similarity with reference audios are limited. In contrast, the output audios of the IT model in Figure~\ref{fig:style_transfer_1.c} are with similar lengths coinciding with texts, which sound natural and similar to the reference. The results indicate that disentangled style embeddings ensure the success of non-parallel style transfer.






\subsubsection{Style control}\label{sec:style_control_1}

\begin{figure}[th]
  \centering
  \subfigure[Reference audios]{
  \label{fig:style_control_1.a}
  \begin{minipage}[b]{1\linewidth}
    \includegraphics[width=0.19\linewidth, height=0.4in]{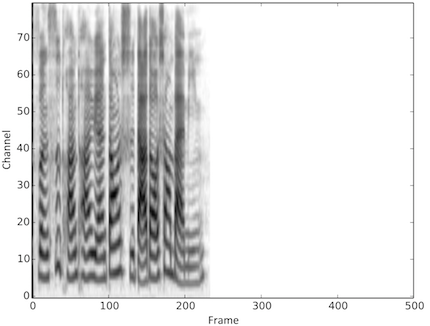}
    \hspace{0.59\linewidth}
    \includegraphics[width=0.19\linewidth, height=0.4in]{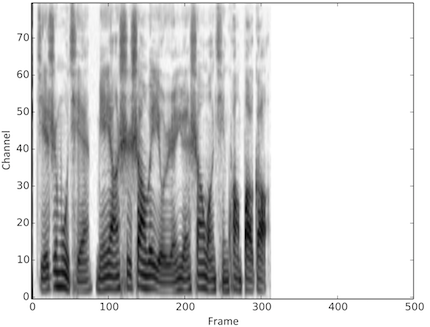}
  \end{minipage}}
  \subfigure[Output audios]{
  \label{fig:style_control_1.b}
  \begin{minipage}[b]{1\linewidth}

    \includegraphics[width=0.19\linewidth, height=0.4in]{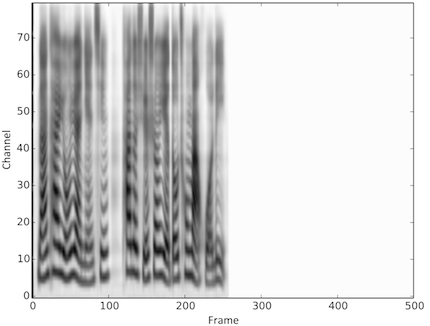}
    \includegraphics[width=0.19\linewidth, height=0.4in]{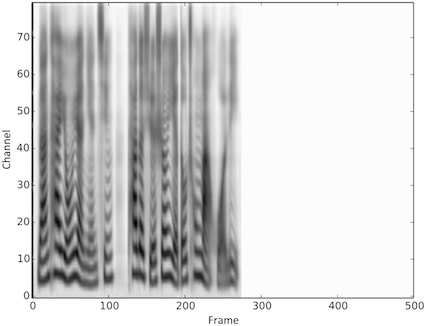}
    \includegraphics[width=0.19\linewidth, height=0.4in]{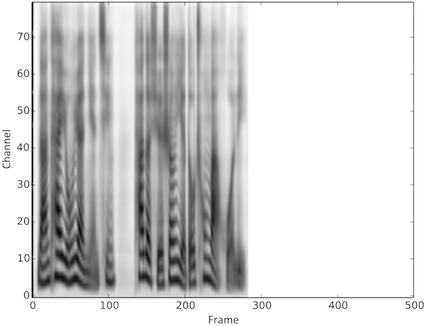}
    \includegraphics[width=0.19\linewidth, height=0.4in]{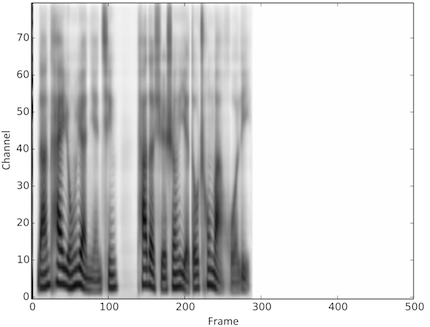}
    \includegraphics[width=0.19\linewidth, height=0.4in]{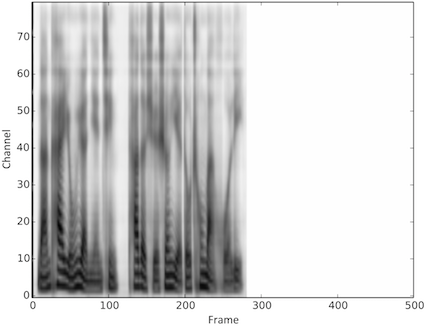}\vspace{0pt}

    \includegraphics[width=0.19\linewidth, height=0.4in]{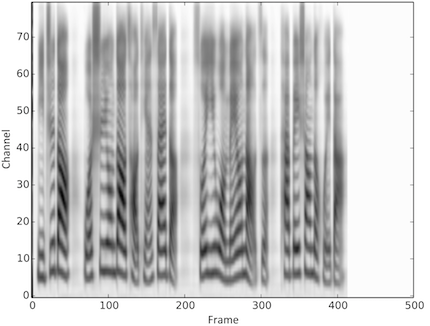}
    \includegraphics[width=0.19\linewidth, height=0.4in]{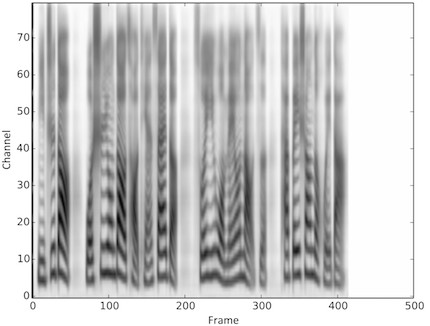}
    \includegraphics[width=0.19\linewidth, height=0.4in]{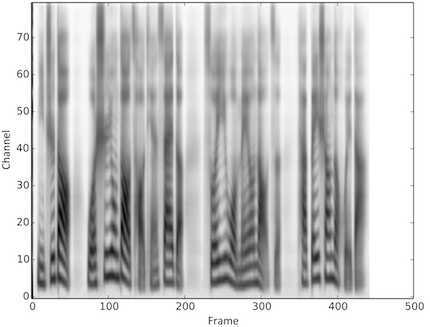}
    \includegraphics[width=0.19\linewidth, height=0.4in]{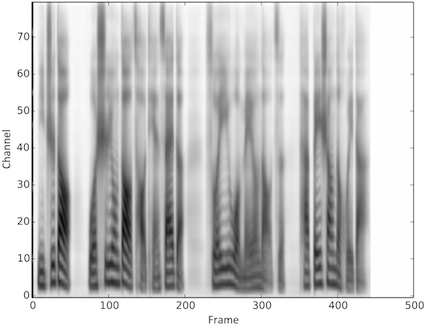}
    \includegraphics[width=0.19\linewidth, height=0.4in]{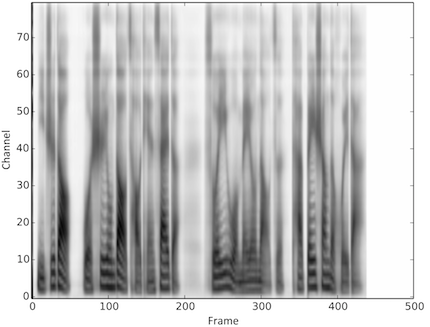}\vspace{0pt}

  \end{minipage}}
  \caption{Style control by linear interpolation. (b) Left to right: referencing style embeddings interpolating from A to B with $\alpha$ set to 0.0, 0.25, 0.5, 0.75, 1.0. Top to bottom: Conditioning on Chinese sentences with 16 tokens, 24 tokens}
  \label{fig:style_control_1}
\end{figure}

We randomly chose two audios from a female speaker A and a male speaker B as references and implement linear interpolation introduced in Section~\ref{sec:inference} over style embeddings from A to B. The results are shown in Figure~\ref{fig:style_control_1}.

We see a smooth conversion from female speaker A to male speaker B in speaker related features such as F0 and prosody simultaneously. The results indicate that the style latent space is continuous and can be interpolated freely. Furthermore, style transfer could be seen as special cases ($\alpha = 0$ or $1$) of style control so it was omitted in multi-reference experiments. 


\subsubsection{Few-shot/one-shot speaker conversion}\label{sec:conversion}

Few-shot/one-shot speaker conversion is an important application in multi-speakers TTS. In our models, it is a special case of style transfer, where the reference audios are out of the training set. As we have achieved linear interpolation, we deduce that if the styles of the extra speakers can be represented as linear combinations of the styles of the existing speakers, then one-shot conversion can be done correctly. And we achieved about 20\% accept rate by one-shot. Then we randomly picked 20 utterances from the failed speaker to fine-tuning the model as few-shot. We found that the model performs best when fine-tuning both multi-reference encoder and decoder while keeping the text encoder fixed. We achieved 100\% accept rate by few-shot. Experiments details are presented in our demo page.


\subsection{Multi-references}
In these experiments, we built 2-reference models to control two style classes: speaker and prosody. Models were trained on a 30 hours corpus with 27 speakers and 5 prosodies. Partial details of the corpus is presented in Table~\ref{tab:non-parallel_corpus}.

\begin{table}[th]
  \caption{Partial details of multi-style (speaker-prosody) data corpus. Empty means no data. M = male, F = female.}
  \label{tab:non-parallel_corpus}
  \centering
  \begin{tabular}{cccccc}
    \toprule
    \textbf{speaker}  & \textbf{news} & \textbf{story} & \textbf{radio} & \textbf{poetry} & \textbf{call-center}  \\
    \midrule
    M2      &       &       &       &402    &       \\
    M5      &423    &440    &453    &388    &       \\
    F4      &       &       &       &       &421    \\
    F17     &1987   &       &467    &376    &464    \\
    \midrule
    Total   &3613   &7101   &3812   &2893   &3013   \\
    \bottomrule
  \end{tabular}
\end{table}

\subsubsection{Style disentangling}

\begin{figure}[th]
  \centering
  \subfigure[Speaker encoder]{
  \label{fig:embedding_visualization_tsne_2.a}
  \begin{minipage}[b]{0.48\linewidth}
    \includegraphics[width=0.47\linewidth, height=0.7in]{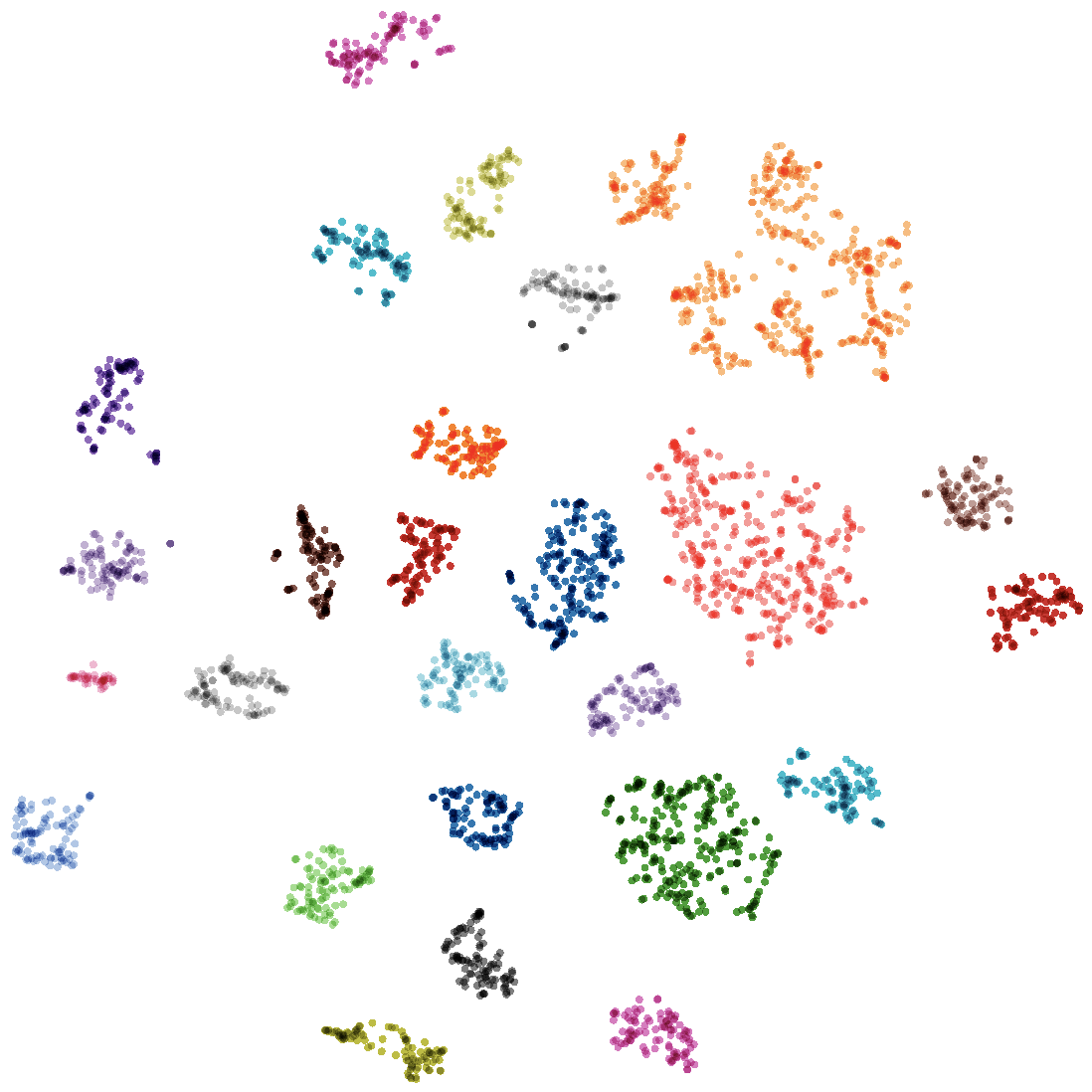}
    \includegraphics[width=0.47\linewidth, height=0.7in]{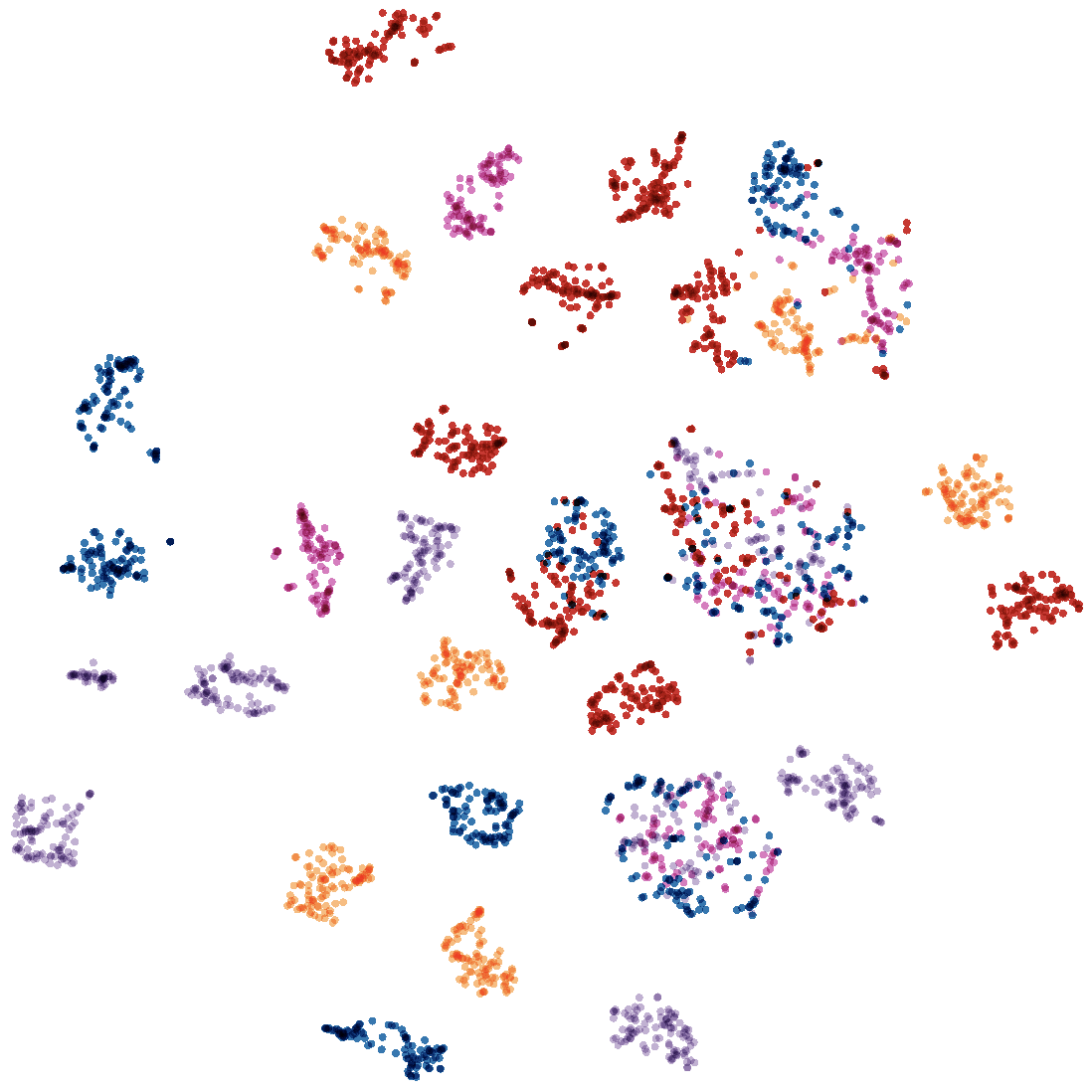}
  \end{minipage}}
  \subfigure[Prosody encoder]{
  \label{fig:embedding_visualization_tsne_2.b}
  \begin{minipage}[b]{0.48\linewidth}
    \includegraphics[width=0.47\linewidth, height=0.7in]{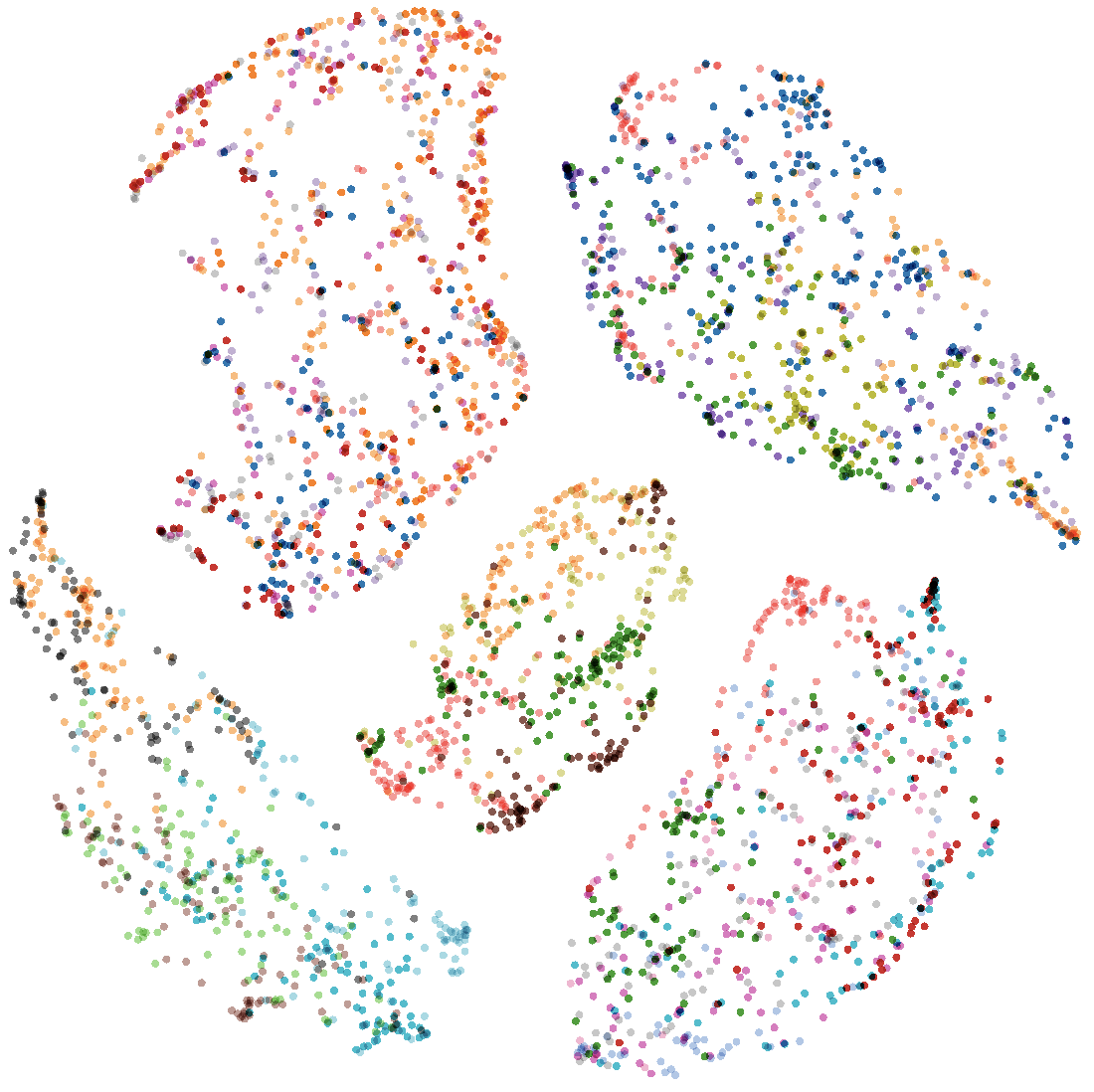}
    \includegraphics[width=0.47\linewidth, height=0.7in]{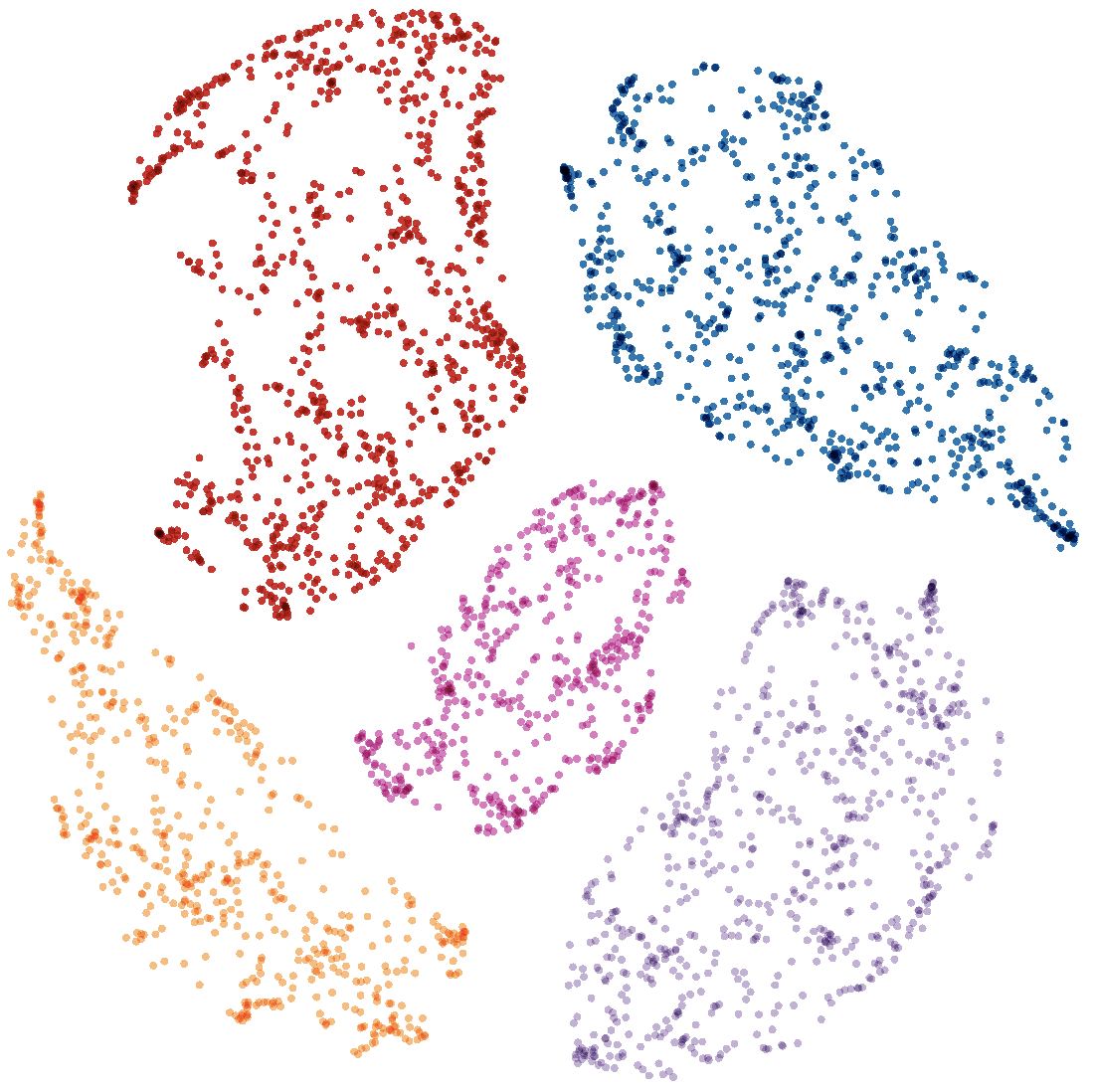}
  \end{minipage}}
  \caption{Visualize style embeddings of two sub-encoders. In each sub-figure, the left picture is colored by speakers and the right picture is colored by prosodies.}
  \label{fig:embedding_visualization_tsne_2}
\end{figure}

We randomly select 100 utterances from each prosody of each speaker, and visualize their style embeddings by t-SNE. Figure~\ref{fig:embedding_visualization_tsne_2} shows the results. 

We see the embeddings from speaker encoder in Figure~\ref{fig:embedding_visualization_tsne_2.a} are clustered only by speaker, while the embeddings from prosody encoder in Figure~\ref{fig:embedding_visualization_tsne_2.b} are clustered only by prosody. The results prove that the speaker and prosody styles are disentangled well into designated encoders, so that we can control these styles independently.

\subsubsection{Style control}

\begin{figure}[th]
  \centering
  \begin{minipage}[t]{1\linewidth}
    \includegraphics[width=0.32\linewidth, height=0.4in]{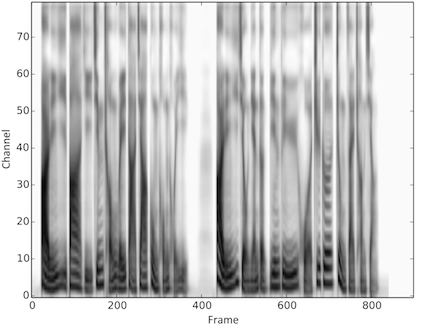}
    \includegraphics[width=0.32\linewidth, height=0.4in]{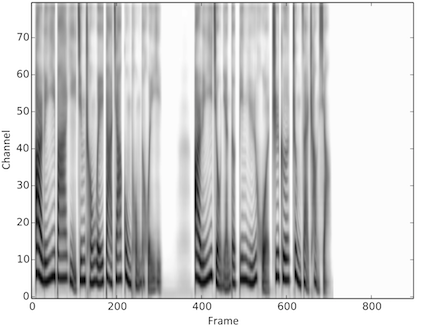}
    \includegraphics[width=0.32\linewidth, height=0.4in]{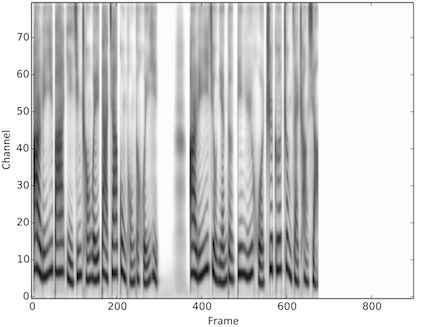}\vspace{0pt}

    \includegraphics[width=0.32\linewidth, height=0.4in]{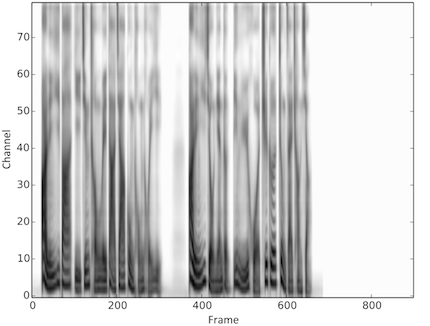}
    \includegraphics[width=0.32\linewidth, height=0.4in]{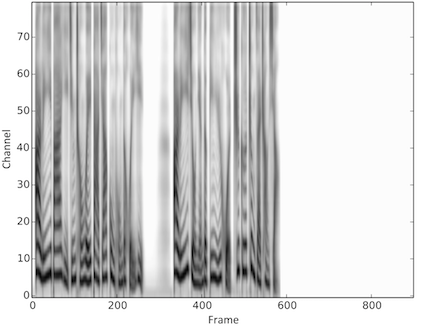}
    \includegraphics[width=0.32\linewidth, height=0.4in]{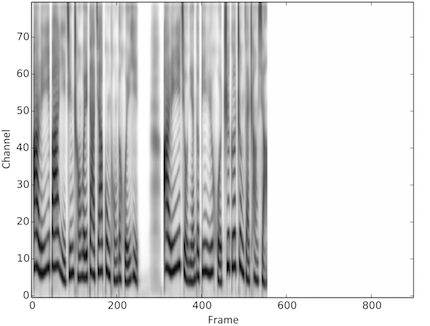}\vspace{0pt}

    \includegraphics[width=0.32\linewidth, height=0.4in]{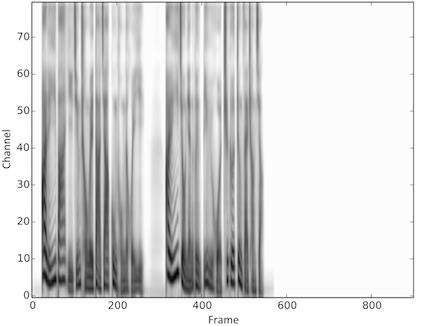}
    \includegraphics[width=0.32\linewidth, height=0.4in]{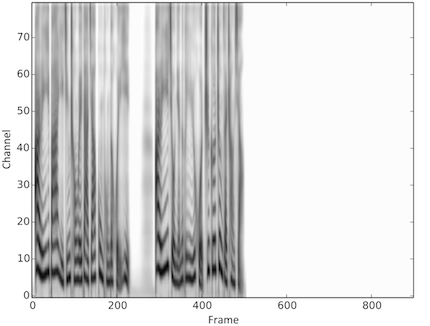}
    \includegraphics[width=0.32\linewidth, height=0.4in]{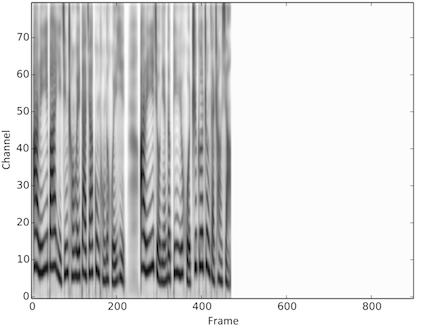}
  \end{minipage}
  \caption{multi-style control over non-parallel data. M2 to F4 from left to right, poetry to callcenter from top to bottom. $\alpha$ of both speaker and prosody is set to 0.0, 0.5, 1.0.}
  \label{fig:style_control_2}
\end{figure}

We conducted style control over both parallel and non-parallel speakers. Here ``non-parallel'' means the speakers have no common target prosody styles, vice versa. Figure~\ref{fig:style_control_2} shows the non-parallel results over M2-poetry and F4-callcenter. The text contents are fixed for better distinguishing the styles.



We see gradual and independent changes of F0 from male to female and audio lengths from poetry to callcenter. The results indicate that different style latent spaces are independent from each other, while each style latent space is continuous, which ensures the success of multi-dimensional linear interpolation.


Furthermore, random sampling and few-shot/one-shot conversion mentioned in Section~\ref{sec:inference} and~\ref{sec:conversion} can also be applied in any sub-encoders of the multi-reference models to synthesize more expressive and diverse voices.

\section{Conclusions}\label{sec:conclusions}

In this paper, we introduced multi-reference encoder to Tacotron and proposed intercross training technique. We prove that our model could disentangle different speech style classes into corresponding reference encoders and could control them independently. We showed that the proposed model could accomplish style disentangling, control, transfer, and other style related tasks in a flexible, interpretable and robust manner.


\bibliographystyle{IEEEtran}



\end{document}